# On Some Equivalence Relations Between Incidence Calculus and Dempster-Shafer Theory of Evidence

Flávio S. Corrêa da Silva
Alan Bundy

Department of Artificial Intelligence,
University of Edinburgh
80 South Bridge, Edinburgh, Scotland EH1 1HN

email: fcs@aipna.ed.ac.uk    bundy@aipna.ed.ac.uk

## Abstract

Incidence Calculus and Dempster-Shafer Theory of Evidence are both theories to describe agents' degrees of belief in propositions, thus being appropriate to represent uncertainty in reasoning systems.

This paper presents a straightforward equivalence proof between some special cases of these theories.

## 1 Introduction

Incidence Calculus [1,2,6] and Dempster-Shafer Theory of Evidence [3,5,8] are two alternative theories to represent uncertain knowledge in reasoning systems. They present a series of similarities:

1. Both of them have been proposed as extensions of the Bayesian approach.

2. Both of them have proposed interval-based probability extensions.

3. Both of them have considered a specific kind of uncertainty, the probability assignment on possible worlds [4].

The main difference between them is that Dempster-Shafer Theory of Evidence extends Bayesian Theory by allowing *possible worlds with undefined probability measures*, whereas Incidence Calculus does so by allowing *propositions with undefined truth assignments* on possible worlds.

In this paper we present some equivalence proofs between these theories, when applied over finite propositional languages. This is made possible due to the results presented by Fagin and Halpern in [3], and the proofs are achievable through the use of a reformulated version of Fagin and Halpern's *probability structures*.

## 2 Preliminary Definitions

### 2.1 Probability Space

Following [3], a *probability space* is a tuple $(S, \chi, \mu)$, where $S$ is a sample space, $\chi$ is a $\sigma$-algebra of $S$[1], and $\mu$ is a probability measure of $\chi$:

$$\mu : \chi \to [0, 1] \text{ such that}$$

1. $\mu(S) = 1$

2. $\mu(\cup_1^\infty X_i) = \sum_1^\infty \mu(X_i)$, $X_i \in \chi$ pairwise disjoint

   (If $\chi$ is finite, this last property turns to

   $\mu(X_1 \cup X_2) = \mu(X_1) + \mu(X_2)$, $X_1 \cap X_2 = \emptyset$, $X_1, X_2 \in \chi$

   and allows us to derive the following corollaries:

   (a) $\mu(\neg X) = 1 - \mu(X)$

---

[1]that is, $\chi$ is a set of subsets of $S$ such that $S \in \chi$ and $\chi$ is closed under complementation and countable union

(b) $\mu(\emptyset) = 0$,

where $\emptyset$ stands for the empty set)

The probability measure is defined only on $\chi$, that is a subset of the power set $2^S$ of $S$. It can be extended on $2^S$ in a standard way, by defining the inner measure induced by $\mu$ ($\mu_*$):

$$\mu_* : 2^S \to [0, 1]$$

$$\mu_*(A) = sup\{\mu(X) : X \subseteq A, X \in \chi\}, A \in 2^S$$

that, due to the codomain of the measure function and property 2., can be rewritten, for finite $\sigma$-algebras, as:

$$\mu_*(A) = \mu(\bigcup X : X \subseteq A, X \in \chi)$$

Given a $\sigma$-algebra $\chi$, the subset $\chi' \subseteq \chi$ is called a *basis* of $\chi$ iff all members of $\chi'$ are disjoint and nonempty and $\chi$ consists precisely of countable unions of members of $\chi'$. It is provable that if $\chi$ is finite then it has a basis [3].

## 2.2 Propositional Language

The language we are going to consider is a finite propositional language. It can be characterised by a finite set of primitive propositions $\Phi = \{p_1, ..., p_n\}$ and its closure under the application of the Boolean operators $\wedge$ and $\neg$. The primitive propositions in $\Phi$ do not necessarily describe mutually exclusive events. In order to have mutually exclusive events, $\Phi$ can be restated as the set $At = \{\delta_1, ..., \delta_{2^n}\}$, where $\delta_i = p'_1 \wedge ... \wedge p'_n$ and $p'_j = p_j$ or $p'_j = \neg p_j$. If we define the operator $\vee$ in terms of $\wedge$ and $\neg$ as usual, then we can associate subsets $\varphi$ of $At$ with formulae in the propositional language generated by $\Phi$, consisted by the disjunction of the elements of $\varphi$. $2^{At}$ represents all the formulae generated by $\Phi$.

Fagin and Halpern [3] assumed that all the formulae generated by $\Phi$ had an interpretation, i.e., allowed a truth assignment. We will assume that a $\sigma$-algebra $\psi$ of $At$ has interpretations. Since $2^{At}$ is a $\sigma$-algebra of $At$, our assumption includes Fagin and Halpern's one.

Since $\Phi$ is finite, a $\sigma$-algebra $\psi \subseteq 2^{At}$ of $At$ will be a set such that:

1. $\emptyset \in \psi$
2. $At \in \psi$
3. $\varphi \in \psi \Leftrightarrow \neg\varphi \in \psi$
4. $\varphi_1, \varphi_2 \in \psi \Rightarrow (\varphi_1 \wedge \varphi_2) \in \psi$
5. $\varphi_1, \varphi_2 \in \psi \Rightarrow (\varphi_1 \vee \varphi_2) \in \psi$

## 3 Probability Structure

In [3] a *probability structure* is defined as a tuple $(S, \chi, \mu, \pi)$, associated with a finite propositional language $At$. $(S, \chi, \mu)$ is a probability space and $\pi$ is a truth-assignment mapping, that associates with each $s \in S$ a mapping $\pi(s) : \Phi \to \{true, false\}$. Intuitively, each $s$ corresponds to a possible world. Alternatively, we define an incidence mapping $i$ as a specific formulation of the inverse mapping of $\pi$, that is:

$$i : \psi \to 2^S$$

with the following properties:

1. $i(\varphi) = \{s \in S : \varphi \text{ is true in } s\}$
2. $i(\emptyset) = \emptyset$
3. $i(At) = S$
4. $i(\neg\varphi) = S - i(\varphi)$
5. $i(\varphi_1 \wedge \varphi_2) = i(\varphi_1) \cap i(\varphi_2)$
6. $i(\varphi_1 \vee \varphi_2) = i(\varphi_1) \cup i(\varphi_2)$

Thus, we restate a *probability structure* as a tuple $(S, \chi, \mu, At, \psi, i)$, where $(S, \chi, \mu)$ is a probability space, $At$ defines a propositional language, $\psi$ is a $\sigma$-algebra of $At$ and $i$ is an incidence mapping.

## 4 Dempster-Shafer Structure

Given a set of mutually exclusive and exhaustive atomic events, Dempster-Shafer Theory of Evidence provides a way to attach degrees of belief on these events by defining *belief* and *plausibility* functions. We can formalise these concepts as tuples $(\Theta, bel)$, where $\Theta$ is the set of atomic events and $bel$ is the *belief function over* $\Theta$, that is:

$$bel : 2^\Theta \to [0,1] \text{ such that}$$

1. $bel(\emptyset) = 0$
2. $bel(\Theta) = 1$
3. $bel(\bigcup_1^k A_i) \geq$
   $\sum_{I \subseteq \{1,...,k\}, I \neq \emptyset}(-1)^{|I|+1} bel(\bigcap_{i \in I} A_i)$,
   where $|I|$ stands for the cardinality of $I$

Observe that the right-hand side of property .3. corresponds to the inclusion-exclusion rule for probabilities [3,7].

The associated *plausibility function over* $\Theta$ derives from the definition of the belief function:

$$plb : 2^\Theta \to [0,1]$$

$$plb(A) = 1 - bel(\neg A)$$

(that is, $plb(\emptyset) = 1$, $plb(\Theta) = 0$,
$plb(\bigcup_1^k A_i) \leq$
$1 - \sum_{I \subseteq \{1,...,k\}, I \neq \emptyset}(-1)^{|I|+1} bel(\bigcap_{i \in I} \neg A_i)$

Thus, to each event $A$ we can associate a subset of $\Theta$ and attach an interval $[bel_i(A), plb_s(A)]$ to which the probability of $A$ belongs, where $bel_i$ is the *inf* of $bel$ and $plb_s$ is the *sup* of $plb$.

In [3], a specific interpretation for Dempster-Shafer Theory of Evidence is provided. The set $\Theta$ is interpreted as a set of possible worlds, to which the truth-evaluation of a finite propositional language is associated. In that case, it is proved that to any tuple $(\Theta, bel)$ there is another tuple $(\Theta', bel')$, in which $\Theta'$ is finite and the evaluations of $bel'$ and $bel$ are equal for any formula in the associated propositional language. It is also proved that to any tuple $(\Theta, bel)$ in which $\Theta$ is finite there is a corresponding probability structure $(S, \chi, \mu, At, 2^{At}, i)$ in which the evaluations of $bel$ and $\mu_*$ are equal for any formula, provided that the propositional language is the same. It is worth observing that $S$ is also finite, in that case.

Thus, if we restrict our attention to Fagin and Halpern's interpretation and finite propositional languages, we can take *Dempster-Shafer structures* as probability structures $(S, \chi, \mu, At, 2^{At}, i)$, in which $bel$ is the inner measure induced by $\mu$ and $S$ is finite.

We define *Total Dempster-Shafer structures* as tuples $(S, \chi, \mu, At, 2^{At}, i)$ in which the image of the incidence mapping contains the $\sigma$-algebra of the probability structure, that is, for all $X \in \chi$ there is a $\varphi \in 2^{At}$ such that $i(\varphi) = X$. In other words, Total Dempster-Shafer structures guarantee that all the measurable sets of possible worlds refer to some formula in the propositional language under consideration.

## 5 Incidence Calculus Structure

Incidence Calculus [1,2,6] was proposed as an alternative way to attach and propagate degrees of belief on propositions. Here we have, as before, a set $S$ of possible worlds, a probability measure associated with sets of possible worlds and an incidence mapping from propositions of a propositional language to possible worlds.

Differently from before, though, the probability measure is defined on the whole power set $2^S$ of $S$ - and the incidence mapping is defined only on a generic $\sigma$-algebra of atomic propositions. These concepts are captured by the probability structure $(S, 2^S, \mu, At, \psi, i)$, where the symbols are defined as before.

The inference rules in the Legal Assignment Finder procedure [1,2] extend the incidence mapping on all formulae, by determining the sets of possible worlds that would contain and the ones that would be contained by those in which the formulae are true. That is, the incidence mapping is extended to *lower* and *upper* evaluations as follows (respectively, $i_*$ and $i^*$):

$$i_*(\xi) = \bigcup[i(\varphi) : \varphi \subseteq \xi, \varphi \in \psi]$$

$$i^*(\xi) = S - \bigcup[i(\varphi) : \varphi \subseteq \neg\xi, \varphi \in \psi] = S - i_*(\xi).$$

$$\xi \in 2^{At}$$

The probability of a formula $\xi$ is then defined, in general, as belonging to the interval $[\mu(i_*(\xi)), \mu(i^*(\xi))]$.

# 6 Equivalence Relations Between Dempster-Shafer and Incidence Calculus Structures

Two probability structures $(S_1, \chi_1, \mu_1, At_1, \psi_1, i_1)$ and $(S_2, \chi_2, \mu_2, At_2, \psi_2, i_2)$ are said to be *equivalent with respect to a propositional language At iff* $At \subseteq At_1$, $At \subseteq At_2$ and they define the same probability interval for every formula in $2^{At}$.

There is a correspondence between Total Dempster-Shafer structures and Incidence Calculus structures - that is, there is an association of an equivalent Total Dempster-Shafer structure to each Incidence Calculus structure, and vice-versa. Intuitively, it means that (Incidence Calculus) probability boundaries over propositions can be "translated" into (Dempster-Shafer) proability boundaries over possible worlds and vice-versa. Formally, these statements can be presented and proved as below:

**Theorem 6.1**

*For every Incidence Calculus Probability Structure there is an equivalent Total Dempster-Shafer Probability Structure.*

**Proof:** Given any Incidence Calculus structure $(S, 2^S, \mu, At, \psi, i)$, we define a Total Dempster-Shafer structure $(At, \psi, \mu_{ds}, At, 2^{At}, i_{ds})$, where $\mu_{ds}(\varphi) = \mu(i(\varphi))$, and $i_{ds}$ is an identity function.

Then we have that

$\mu(i_*(\xi)) = \mu(\bigcup \varphi : \varphi \subseteq \xi, \varphi \in \psi) = \mu_{ds}(\bigcup \varphi : \varphi \subseteq i_{ds}(\xi), \varphi \in \psi) = bel(\xi)$

$\mu(i^*(\xi)) = \mu(S - \bigcup \varphi : \varphi \subseteq \neg\xi, \varphi \in \psi) = \mu(S) - \mu(\bigcup \varphi : \varphi \subseteq \neg\xi, \varphi \in \psi) =$
$= 1 - \mu(i_*(\neg\xi)) = 1 - bel(\neg\xi) = plb(\xi)$.

QED

**Theorem 6.2**

*For every Total Dempster-Shafer Probability Structure there is an equivalent Incidence Calculus Probability Structure.*

**Proof:** Given a Total Dempster-Shafer structure $(S, \chi, \mu, At, 2^{At}, i)$ we define an Incidence Calculus structure $(S_{ic}, 2^{S_{ic}}, \mu_{ic}, At, \psi_{ic}, i_{ic})$, where $S_{ic} = \chi'$, $\mu_{ic} = \mu$, $\psi'_{ic} = \{\{\delta_i\} : i(\delta_i) \in S_{ic}, \delta_i \in At\}$, and $i_{ic}(\varphi) = i(\varphi), \varphi \in \psi_{ic}$.

Then we have that

$bel(\xi) = \mu(\bigcup X : X \subseteq i(\xi), X \in \chi) = \mu(\bigcup(i(\varphi)) : i(\varphi) \subseteq i(\xi), i(\varphi) \in \chi) =$
$= \mu(\bigcup(i(\varphi)) : \varphi \subseteq \xi, i(\varphi) \in \chi) = \mu(\bigcup(i(\varphi)) : \varphi \subseteq \xi, \varphi \in \psi_{ic}) =$
$= \mu(\bigcup(i_{ic}(\varphi)) : \varphi \subseteq \xi, \varphi \in \psi_{ic}) = \mu_{ic}(\bigcup(i_{ic}(\varphi)) : \varphi \subseteq \xi, \varphi \in \psi_{ic}) =$
$= \mu_{ic}(i_{ic}(\bigcup \varphi : \varphi \subseteq \xi, \varphi \in \psi_{ic})) = \mu_{ic}(i_*(\xi))$

$plb(\xi) = 1 - bel(\neg\xi) = 1 - \mu_{ic}(i_*(\neg\xi)) = \mu_{ic}(S) - \mu_{ic}(i_*(\neg\xi)) =$
$= \mu_{ic}(S - i_*(\neg\xi)) = \mu_{ic}(i^*(\neg\xi))$.

QED

# 7 Example

As an example, one problem is solved by using equivalent structures belonging to each theory. The problem is the one presented in [3] - example 2.2, and stated here as follows:

> A person has four coats: two are blue and single-breasted, one is grey and double-breasted and one is grey and single-breasted. To choose which color of coat this person is going to wear, one tosses a (fair) coin. Once the colour is chosen, to choose which specific coat to wear the person uses a mysterious nondeterministic procedure which we don't know anything about. What is the probability of the person wearing a single-breasted coat?

## 7.1 Dempster-Shafer Solution

Let $S_{ds} = \{s_1, s_2, s_3, s_4\}$, where $s_1$ and $s_2$ correspond to the wearing of blue coats, $s_3$ corresponds to the wearing of the grey single-breasted coat and $s_4$ corresponds to the wearing of the grey double-breasted one.

Let $\Phi = \{g, d\}$, corresponding respectively to "the coat is grey" and "the coat is double-breasted".

Then we have

$At = \{g \wedge d, \neg g \wedge d, \neg g \wedge \neg d, g \wedge \neg d\},$

$\chi_{ds} = \{\{s_1, s_2\}, \{s_3, s_4\}, \emptyset, S_{ds}\},$

$\mu(\{s_1, s_2\}) = \mu(\{s_3, s_4\}) = 0.5, \mu(\{\emptyset\}) = 0, \mu(\{S\}) = 1.$

Observe that this structure is total.

The desired answer,

$\mu(i(\neg d)) = \mu(\{s_1, s_2, s_3\}),$

is undefined. The lower and upper bounds for this value are:

$bel(\neg d) = \mu_*(i(\neg d)) = \mu_*(\{s_1, s_2, s_3\}) = \mu(\{s_1, s_2\}) = 0.5,$

$plb(\neg d) = 1 - \mu_*(i(d)) = 1 - \mu_*(\{s_4\}) = 1 - \mu(\emptyset) = 1.$

It means that, although we have no means to evaluate the probability of the person to wear a single-breasted coat, we know that it is not smaller than 0.5.

### 7.2 Incidence Calculus Solution

To solve the same problem using Incidence Calculus, instead of constructing a set of possible worlds - some of which being nonmeasurable - with each one corresponding to one possible situation, we construct a set of possible worlds with each one corresponding to one *measurable* situation, that may be a set of subsets of the formerly considered situations.

For example, let's construct an Incidence Calculus structure by applying the procedure pointed out in Theorem 6.2 on the formulation of section 7.1:

Let $S_{ic} = \{w_1, w_2\}$, where $w_1$ and $w_2$ correspond to the possible worlds in which the blue and grey coats are worn, respectively. This is the basis of $\chi_{ds}$ ($w_1 = \{s_1, s_2\}, w_2 = \{s_3, s_4\}$).

Let $\Phi$ and $At$ be as before. Now,

$2^{S_{ic}} = \{\emptyset, \{w_1\}, \{w_2\}, S_{ic}\},$

$\psi'_{ic} = \{\emptyset, \neg g \wedge d, \neg g \wedge \neg d, (g \wedge \neg d) \vee (g \wedge d), (\neg g \wedge \neg d) \vee (\neg g \wedge d), (\neg g \wedge d) \vee (g \wedge \neg d) \vee (g \wedge d), (\neg g \wedge \neg d) \vee (g \wedge \neg d) \vee (g \wedge d), At\},$

and $\mu(\{w_1\}) = \mu(\{w_2\}) = 0.5, \mu(\{\emptyset\}) = 0, \mu(\{S\}) = 1.$

The desired answer, $\mu(i(\neg d)) = \mu(i((\neg g \wedge \neg d) \vee (g \wedge \neg d)))$ is undefined because $i((\neg g \wedge \neg d) \vee (g \wedge \neg d))$ is undefined. We have that:

$\mu(i_*(\neg d)) = \mu(i(\neg g \wedge \neg d)) = \mu(\{s_1\}) = 0.5,$
$\mu(i^*(\neg d)) = 1 - \mu(i(\neg g \wedge d)) = 1 - \mu(\emptyset) = 1$

and we find the same values as before.

## 8 Conclusion and Further Work

We have shown that there is an equivalence relation between Total Dempster-Shafer structures and Incidence Calculus structures for finite propositional languages.

Apart from the proof of equivalence itself, this result makes it possible to restate problems with incomplete information about probabilities on possible worlds as problems with partial truth-assignment functions on formulae and vice-versa, what can be useful in a knowledge-base design process for providing different views of one problem, thus helping the expert to explicitate his knowledge.

Further work includes the analysis of the formal relationship between Incidence Calculus as presented here and as in the original papers [1,2] - specifically, we must prove that the reconstruction presented here corresponds with the rules of inference in the Legal Assignment Finder procedure [1,2] - and attempting to drop the restrictions of finiteness of the propositional language and "totalness" of the Dempster-Shafer structure.

**Acknowledgements:** The authors are grateful for the several comments and suggestions from Paul Chung, Joe Halpern and Dave Robertson.

The first author is a PhD student at the University of Edinburgh, and is being supported by a scholarship from Conselho Nacional de Desenvolvimento Científico e Tecnológico - CNPq - Brazil, grant nr. 203004-89.2.